\documentclass{article}
\usepackage{ijcai25}

\usepackage{pifont}
\usepackage{multirow}
\usepackage{wrapfig}
\usepackage{amssymb}
\usepackage{times}
\usepackage{soul}
\usepackage{url}
\usepackage[hidelinks]{hyperref}
\usepackage[utf8]{inputenc}
\usepackage[small]{caption}
\usepackage{graphicx}
\usepackage{amsmath}
\usepackage{amsthm}
\usepackage{booktabs}
\usepackage{algorithm}
\usepackage{algorithmic}
\usepackage[switch]{lineno}
\usepackage{xcolor}
\usepackage{xspace}

\newcommand{\ie}{i.e.\xspace}
\newcommand{\ours}{\textbf{\texttt{Latte}}}

\title{\ours: Transfering LLMs' Latent-level Knowledge for Few-shot Tabular Learning}

\author{
Ruxue Shi\thanks{Equal contribution.}
\and
Hengrui Gu\footnotemark[1]\and
Hangting Ye\and
Yiwei Dai\and
Xu Shen\And
Xin Wang\thanks{Corresponding author.}
\\
\affiliations
Jilin University, Changchun, China\\
\emails
{
\{shirx24, guhr22, yeht2118, daiyw23, shenxu23\}@mails.jlu.edu.cn,
xinwang@jlu.edu.cn
}
}

\begin{document}

\maketitle

\def\myVar{Latte}

\begin{abstract}

Few-shot tabular learning, in which machine learning models are trained with a limited amount of labeled data, provides a cost-effective approach to addressing real-world challenges. The advent of Large Language Models (LLMs) has sparked interest in leveraging their pre-trained knowledge for few-shot tabular learning. Despite promising results, existing approaches either rely on test-time knowledge extraction, which introduces undesirable latency, or text-level knowledge, which leads to unreliable feature engineering. To overcome these limitations, we propose \ours, a training-time knowledge extraction framework that transfers the latent prior knowledge within LLMs to optimize a more generalized downstream model. \ours~enables general knowledge-guided downstream tabular learning, facilitating the weighted fusion of information across different feature values while reducing the risk of overfitting to limited labeled data. Furthermore, \ours~is compatible with existing unsupervised pre-training paradigms and effectively utilizes available unlabeled samples to overcome the performance limitations imposed by an extremely small labeled dataset.  Extensive experiments on various few-shot tabular learning benchmarks demonstrate the superior performance of \ours, establishing it as a state-of-the-art approach in this domain\footnote{Our code is available at \url{https://github.com/ruxueshi/Latte.git}}.

\end{abstract}

\section{Introduction}

Given the high cost of sample annotation, both in terms of financial and temporal resources, learning from a limited number of labeled samples presents a more cost-effective approach for adapting machine learning models to practical deployment~\cite{snell2017prototypical,wang2020generalizing,oreshkin2018tadam}. This scenario, commonly referred to as \textbf{few-shot learning}, has recently attracted significant research attention over various domains, including computer vision (CV)~\cite{chen2019closer} and tabular learning~\cite{nam2023stunt}. However, without adequate supervisory signals, traditional supervised learning in this situation cannot effectively yield desirably parameterized models, as its performance heavily depends on the statistical convergence of a large number of labeled samples. This challenge becomes more intractable in the context of tabular data, owing to the inherent scarcity of labeled tabular data~\cite{liud2r2} in many real-world applications, such as fraud detection~\cite{frauddetection} and disease diagnosis~\cite{diseasediag}.

With the rapid advancement of Large Language Models (LLMs), several pioneering studies have explored their potential in few-shot tabular learning and achieved promising results: The extensive knowledge encoded during pre-training makes LLMs a natural choice for analyzing tabular data, as it provides a rich prior understanding of the relationships between diverse tabular features and the task objective. For instance, knowledge regarding the increased risk of certain diseases in older individuals can be helpful in identifying truly predictive tabular features in disease prediction tasks~\cite{featllm}. However, most existing LLM-based methods~\cite{slack2023tablet,hegselmann2023tabllm} adopt test-time knowledge extraction strategies, where each test sample is directly fed into a pre-trained or fine-tuned LLM as a query to obtain its final prediction. This approach necessitates at least one LLM inference per sample, leading to undesirable response latency and increased computational costs. 

To solve above limitations, FeatLLM~\cite{featllm} is first proposed by designing a training-time extraction strategy. Specifically, in this framework, the LLM functions as a ``feature engineer'' to reduce the inherent redundancy in the original tabular features: Before each tabular learning task begins, FeatLLM instructs the ``feature engineer'' to analyze the metadata (including task and feature descriptions) and generate insightful rules for each answer class. These rules guide the subsequent feature engineering, i.e., transforming heterogeneous column features of a tabular dataset into unified binary features. This approach reduces the reliance on complex model architectures, thereby alleviating the urgent need for a large number of labeled samples in few-shot settings. 

However, FeatLLM still faces several bottlenecks that limit its performance potential: \ding{182} \textbf{Unreliable knowledge extraction}: FeatLLM prompts LLMs to generate textual rules as the criteria for feature engineering. However, this kind of text-level knowledge, i.e. textual responses auto-regressively generated by LLMs, often suffers from hallucinations and might include plausible yet factually wrong knowledge~\cite{hallucination,hallucination2}, thereby leading to unreliable feature engineering. In addition,~\cite{infor1,infor2} have demonstrated that the latent-level knowledge, i.e. hidden states directly hooked from LLMs, is more informative and discriminative than the text-level knowledge due to the randomness and information loss during the next-token sampling; \ding{183} \textbf{Overlook of unlabeled tabular data}: Constructing pseudo-supervision (e.g., by data clustering) based on unlabeled data has been proven to be an effective practice to facilitate few-shot tabular learning~\cite{nam2023stunt}. However, the tabular learning framework of FeatLLM cannot make use of these cluster pseudo-labels, as its feature engineering process is designed exclusively for ground-truth classes, implying its performance upper-bound is strictly constrained by the extremely limited labeled samples.

\textbf{Contributions:} To accommodate the above bottlenecks, we propose a training-time knowledge extraction framework to transfer LLMs' \underline{\textbf{Lat}}en\underline{\textbf{t}}-level Knowledg\underline{\textbf{e}} for few-shot tabular learning (\ours). Unlike FeatLLM, which treats the LLM as a ``feature engineer'', \ours~positions the LLM as a ``teacher'' to instruct the training process of the downstream tabular learning model. Specifically, for a tabular learning task, \ours~inputs its metadata (i.e., task and feature descriptions) into the LLM and retrieves the hidden states produced by the final transformer layer as task-relevant knowledge. To distill this latent-level knowledge into downstream training, we propose two key components: 1) a semantic-aware tabular encoder that integrates feature semantics into the representation of feature values; and 2) a knowledge adapter that takes this latent-level knowledge as the guideline to weightedly fuse information from different feature values, thereby constructing more predictive row embeddings. Together, these two components enable a general knowledge-guided training process, mitigating the risk of overfitting to limited labeled samples—a persistent challenge in few-shot settings—and thereby promoting convergence toward a more generalized model checkpoint. (\textbf{Bottleneck \ding{182}}). In addition, \ours~includes an additional unsupervised pre-training stage, making full use of the available unlabeled samples to provide a robust parameter initialization for the subsequent knowledge-guided training (\textbf{Bottleneck \ding{183}}).

We conducted extensive experiments to validate \ours~'s effectiveness on few-shot tabular learning tasks. \ours~consistently outperforms all competing methods by a significant margin across various datasets and prediction tasks. Comprehensive ablation studies further highlight the superiority of latent-level knowledge over text-level knowledge, as well as the effectiveness of the individual components in this framework.

\section{Related work}
\subsection{Classical Tabular Prediction}
A wide range of machine learning methods have been developed for tabular data. For modeling linear relationships, logistic regression (LR)~\cite{lavalley2008logistic} is commonly used. In contrast, tree-based models, such as decision trees (DT)~\cite{loh2011classification} and ensemble methods like XGBoost~\cite{chen2016xgboost}, random forests~\cite{breiman2001random}, CatBoost~\cite{prokhorenkova2018catboost}, and LightGBM~\cite{ke2017lightgbm}, are preferred for modeling non-linear relationships. With the rise of deep learning, there has been growing interest in applying neural networks to tabular data. These approaches can be broadly categorized into four types:  (1) Standard Neural Networks: This group includes methods like SNN~\cite{klambauer2017self}, AutoInt~\cite{song2019autoint}, and DCN V2~\cite{wang2021dcn}. (2) Hybrid Methods: These combine decision trees and neural networks for end-to-end training, with notable examples being NODE~\cite{popov2019neural}, GrowNet~\cite{badirli2020gradient}, TabNN~\cite{ke2018tabnn}, and DeepGBM~\cite{ke2019deepgbm}. (3) Transformer-Based Methods: These models, such as TabNet~\cite{arik2021tabnet}, TabTransformer~\cite{huang2020tabtransformer}, and FT Transformer~\cite{gorishniy2021revisiting}, utilize attention mechanisms to learn from tabular data. (4) Representation Learning Methods: These methods leverage self-supervised and semi-supervised learning to extract meaningful representations, with prominent examples including VIME~\cite{yoon2020vime}, SCARF~\cite{bahri2021scarf}, SAINT~\cite{somepalli2021saint}, and Recontab~\cite{chen2023recontab}.

\begin{figure*}[t]
\centering
\includegraphics[width=1\textwidth]{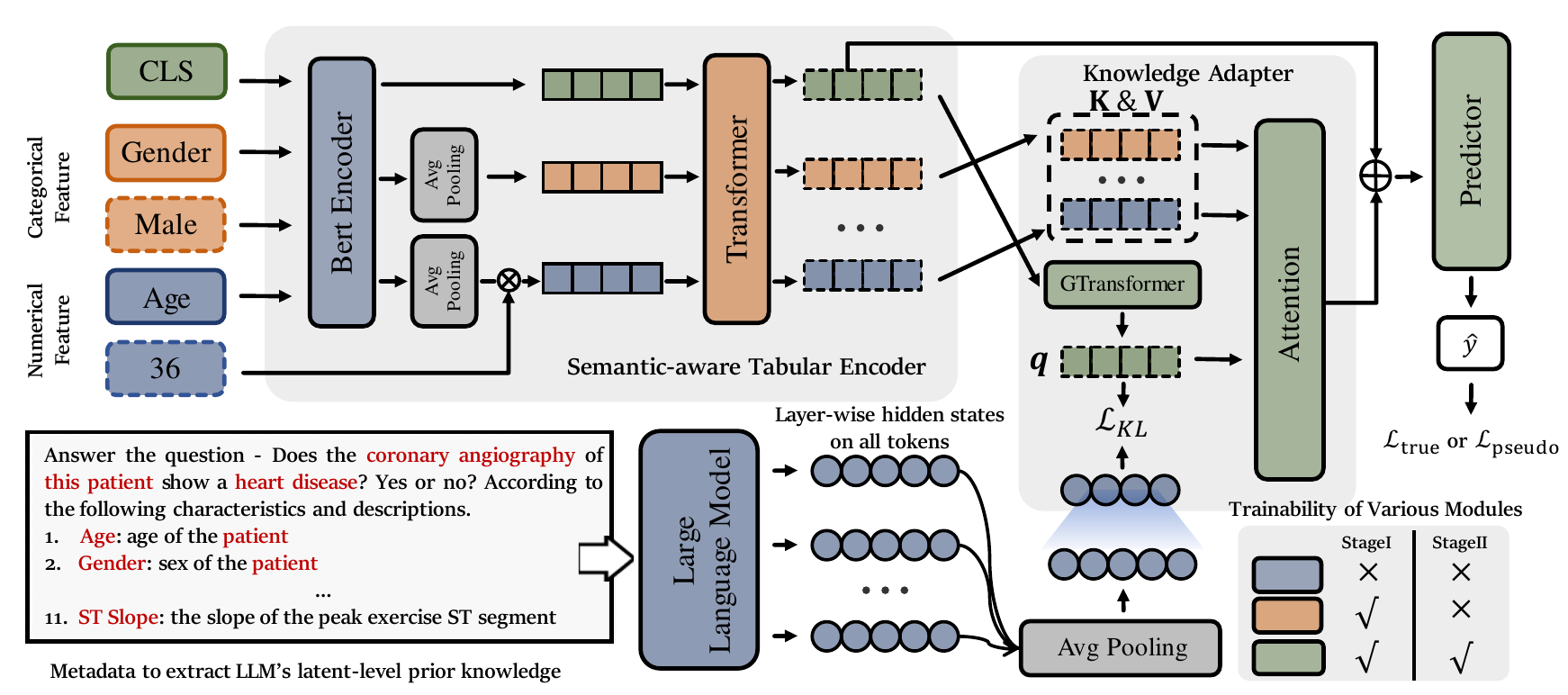}
\caption{\label{figure:method}Overview of our proposed \ours. This framework begins with the extraction of task-relevant knowledge from LLMs. Knowledge adapters are then employed to guide pre-training on unlabeled data, followed by semantic-aware fine-tuning using few-shot labeled samples. Here we refer to the \textbf{unsupervised pre-training} and the \textbf{semantic-aware fine-tuning} as StageI and StageII, respectively.
}
\vspace{-12pt}
\end{figure*}

\subsection{Few-shot Tabular Prediction}
Few-shot tabular prediction holds significant promise in fields like agriculture, industry, and finance, where obtaining and annotating data can be challenging. As a result, this area has garnered increasing attention. However, traditional tabular prediction methods often require large amounts of annotated data and tend to perform poorly in few-shot scenarios. To address these challenges, two main approaches have emerged: (1) TNNs-based methods: These approaches generate tasks from unlabeled data for meta-learning and use a small amount of labeled data to create prototypes for tabular classification tasks~\cite{nam2023stunt}. (2) LLMs-based methods: These rely on the inference capabilities of large language models. For example, the In-context method~\cite{wei2022emergent} directly converts tabular data into text prompts for the LLMs. TABLET~\cite{slack2023tablet} further enhances this by adding task-specific instructions to the prompts, improving the model's ability to reason about tabular. FeatLLM~\cite{han2024large} uses LLMs as a feature engineer to automatically perform feature extraction, which is then used to train specialized classification networks. Alternatively, LLMs can be fine-tuned using a small number of labeled samples, as demonstrated by methods like TabLLM~\cite{hegselmann2023tabllm}. Additionally, GTL~\cite{wen2024supervised} fine-tunes LLMs using a large amount of tabular data to obtain a tabular foundation model.

\section{Method}

In this section, we innovatively propose \ours~to transfer LLMs’ latent-level knowledge for few-shot tabular learning. As shown in  Fig. \ref{figure:method}, \ours~first utilizes a semantic-aware tabular encoder to obtain the representation of feature values integrated with their corresponding semantics (Section \ref{sec:SaTE}). Next, \ours~leverages task metadata to extract task-relevant knowledge from LLMs (Section \ref{sec:LLM}) and transfers this latent-level knowledge into downstream training via a specially designed knowledge adapter (Section \ref{sec:KA}). Finally, we outline the overall downstream training process (Section \ref{sec:MT}).

\subsection{Problem Formulation}

In this section, we formally define the problem settings for few-shot tabular learning: Consider a tabular dataset \( D = \{D_l, D_u, M\} \), where \( D_l \) represents the set of labeled samples,  \( D_u \) the set of unlabeled samples, and \(M\) the corresponding metadata, respectively. Specifically, each sample in the tabular dataset is in this form \((x,y)\), where \(x\) denotes all feature values of a single row, and \(y\) represents its corresponding label. The labeled set \( D_l = \left\{ (x_{l}^{i}, y_{l}^{i}) \right\}_{i=1}^{N_l} \) contains \( N_l \) labeled instances, and the unlabeled set \( D_u = \left\{ x_{u}^{i} \right\}_{i=1}^{N_u} \) consists of \(N_u\) unlabeled instances, with \( N_l \ll N_u \).
Metadata \( M = \{F, T\} \) provides textual descriptions of the task objective and the definitions of features in the dataset, where \( F = \{f_j: d_j\}_{j=1}^{d} \) represents the column names and their descriptions, and \( T \) denotes the task description. Formally, the goal is to train a downstream model\( f_\theta : \mathcal{X} \to \mathcal{Y} \), parameterized by \( \theta \), to map the row feature space \( \mathcal{X} \) to the label space \( \mathcal{Y} \), where \( \mathcal{Y} \) could be \( \{0, 1\}^C \) with \(C\) classes or \( \mathbb{R} \), depending on the specific task (classification or regression). 


\subsection{Semantic-aware Tabular Encoder}
\label{sec:SaTE}
Tabular data presents unique challenges compared to text and images due to its inherent heterogeneity, permutation invariance, and the rich semantic information provided by column names. These characteristics of tabular complicate the encoding process. Traditional feature value encoders struggle with limited labeled samples, making it difficult to infer the semantic meaning of feature names implicitly due to the scarcity of supervisory signals. To address this limitation and better handle the heterogeneity of tabular data, we propose an encoding method that explicitly incorporates feature semantic information into the representation of feature values:
\begin{equation}
h_j =
\begin{cases} 
\text{Average}(\text{BERT}([f_j, x_j])), & \text{for Cat}, \\
\text{Average}(\text{BERT}([f_j])) \times x_j, & \text{for Num},
\end{cases}
\end{equation}
\noindent Here, \(f_j\) denotes the \(j\)-th feature (e.g., ``Gender''), \(x_j\) is its corresponding value (e.g., ``Male''), and \(\times\) represents element-wise multiplication between numerical features and values, because BERT's tokenizer typically splits numerical values into subcomponents, which can hinder its effectiveness for encoding numerical features. This approach generates effective tabular semantic representations for both categorical and numerical features, resulting in a feature set \(H = \{h_j\}_{j=1}^n\), where \(n\) is the total number of features. To obtain a global representation of the sample, we introduce an extra token [CLS] to the feature set \(H\). This token serves as an aggregate representation of the entire sample, capturing the overall context and facilitating downstream tasks.

The interaction between tabular features is crucial for capturing meaningful patterns. For instance, the relationship between weight and blood pressure (BP) can provide valuable insights into cardiovascular health, which is vital for heart disease classification. To model these interactions while preserving the inherent permutation invariance of tabular data, we employ a Transformer architecture without positional embeddings to further refine the quality of feature value embeddings:
\begin{equation}
\hat{H} = \text{Transformer}(H),
\end{equation}
\noindent Here, \(\hat{H} = \{\hat{h}_{\text{CLS}}, \hat{h}_j\}_{j=1}^n\), where  \(\hat{h}_{\text{CLS}}\) and \(\{\hat{h}_j\}_{j=1}^n\) refers to the global representation and the set of tabular feature representations, respectively. These capture the implicit relationships between features, enabling the model to effectively understand and leverage feature interactions.
\subsection{Task-relevant Prior Knowledge Extraction}
\label{sec:LLM}

The above semantic-aware tabular encoder effectively processes heterogeneous tabular data and incorporates its semantic information into the representations of feature values. However, when labeled data is limited, capturing task-specific semantics becomes more challenging. Thanks to large-scale pretraining, LLMs encode substantial general knowledge within their parameters. For a given a tabular learning task, to effectively obtain the task-relevant knowledge within LLMs, we formulate the following knowledge extraction process: Based on the metadata \(M\) of the given tabular dataset \(D\), we elaborately craft the input prompt, describing the task objective and feature semantics (See Fig.~\ref{figure:method} for an example of input prompt), in order to retrieve task-relevant knowledge from the pre-trained LLM. Specifically, after the LLM processes this prompt, we hook the hidden states produced at the last transformer layer of all input tokens and apply average pooling to them. This operation results in a dense vector that encapsulates the LLM's prior understanding of the task, serving as our latent-level knowledge for subsequent knowledge-guided downstream training. The process can be formally described as follows:

\begin{equation}
    [h_1, h_2, ..., h_i, ... ] = \text{LLM}(M),
\end{equation}
\noindent where \(h_i\) represents the last-layer hidden states of the \(i\)-th token after processing the input prompt. We obtain the prompt representation, \ie the latent-level knowledge, by applying average pooling to the hidden states of all tokens:

\begin{equation}
    h_M = \text{Average}([h_1, h_2, ..., h_i, ... ]).
\end{equation}
\noindent where \( \text{Average}(\cdot) \) represents the average pooling operator and \(h_M\) denotes the representation of the input prompt. We regard this dense vector, which contains abundant prior understandings of the given tabular dataset, as the useful latent-level knowledge for subsequent model training.

\subsection{Knowledge Adapter}
\label{sec:KA}
The semantic spaces of LLMs and the BERT-based Semantic-aware Tabular Encoder differ significantly in both dimension and representation, making it challenging to directly integrate the Latent-level knowledge from the LLM into our model. To address this, we design a knowledge adapter that aligns these two semantic spaces and facilitates the transfer of LLM-generated knowledge into our model. We introduce a novel GTransformer to generate the global query vector \( q \):
\begin{equation}
q = \text{GTransformer}(\hat{h}_{CLS}, \{\hat{h}_j\}_{j=1}^n, \{\hat{h}_j\}_{j=1}^n),
\end{equation}
\noindent where both \(\hat{h}_{CLS}\) and \(\{\hat{h}_j\}_{j=1}^n\) are representations encoded by the Semantic-Aware Tabular Encoder. To distill task-related semantic knowledge from the LLM into the global query vector \(q\) and obtain the task-relevant global query vector \(q_{\text{LLM}}\), we apply the following knowledge distillation formula:
\begin{equation}
q_{\text{LLM}} = \text{KL}\left(\textbf{W}_\textbf{0} h_M / \tau, q / \tau \right),
\end{equation}
\noindent where \(\tau\) is the temperature, the matrix \(\textbf{W}_\textbf{0}\) is initialized randomly using Kaiming initialization, which aligns the dimensions of \(h_M\) and \(q\). Meanwhile, through the attention mechanism, we obtain the LLMs semantic knowledge guided task-relevant semantic representation \(h_\text{LLM}\) as:
\begin{equation}
a_i = \text{softmax}\left(\frac{q_\text{LLM} \textbf{K}_i^\top}{\sqrt{d}}\right), 
\end{equation}
\begin{equation}
h_{\text{LLM}} = \sum_{i=0}^{n} a_i\textbf{V}_i,
\end{equation}
\noindent where the \(\hat{H}\) generated by the Semantic-aware Tabular Encoder is directly used as \(\mathbf{K}\) and \(\mathbf{V}\), ensuring that the semantic information of the sample remains unaffected and \(d\) is the feature dimension of \(q_\text{LLM}\). Finally, we combine the general tabular semantics \( \hat{h}_{CLS} \) from the Semantic-aware Tabular Encoder with the task-relevant semantic representation \( h_{\text{LLM}} \) to obtain a high-quality representation \( \hat{h}_{\text{LLM}} \), which effectively integrates both task-relevant and general tabular semantic information:
\begin{equation}
\hat{h}_{\text{LLM}} = \eta \cdot g(h_{\text{LLM}}; \phi) + (1-\eta) \cdot \hat{h}_{CLS},
\end{equation}
\noindent where \(g(\cdot; \phi)\) is a learnable transformation designed to align the semantic spaces of \( h_{\text{LLM}} \) and \( \hat{h}_{CLS} \) and \(\eta\) is a constant that allows for fine-tuning the influence of the LLM’s prior knowledge on the prediction process, ensuring an optimal balance between the model's learned features and the external knowledge introduced.
\subsection{Model Training}
\label{sec:MT}
To fully leverage the available unlabeled samples and ensure robust parameter initialization for subsequent knowledge-guided training, we include an unsupervised pre-training stage, in addition to downstream task fine-tuning. During the pre-training stage, we generate \(N\)-way \(K\)-shot meta-training tasks using unlabeled datasets and perform meta-learning, guided by task-relevant knowledge extracted from LLMs. To obtain the meta-training task, we first generate pseudo-labels \(\tilde{\mathbf{y}}\) for the unlabeled data using the following cluster procedure:
\begin{equation}
\begin{aligned}
\min_{\mathbf{C} \in \mathbb{R}^{d \times k}} \frac{1}{N_u} \sum_{i=1}^{N_u} \min_{\tilde{\mathbf{y}} \in \{0,1\}^k} 
    \left\| (\hat{\mathbf{h}}_{\text{CLS}} \odot \mathbf{m}) - \mathbf{C} \tilde{\mathbf{y}} \right\|_2^2
    & \\
    \text{such that} \quad \tilde{\mathbf{y}}^\top \mathbf{1}_k = 1,
\end{aligned}
\end{equation}
\noindent where \(k\) represents the number of centroids, \(\mathbf{1}_k\) is a vector of ones, \(\mathbf{C}\) is the centroid matrix, and \(\mathbf{m}\) is random noise used to corrupt features, aiding in the learning of robust embeddings for downstream tasks. Next, we randomly select \(k\) samples from each cluster to create an \(N\)-way \(K\)-shot meta-training task. By introducing different noise \textbf{m}, we generate a variety of meta-tasks. Finally, we compute the meta-learning loss as follows:
\begin{equation}
\mathcal{L}_{\text{meta}} = \mathcal{L}_{KL} + \mathcal{L}_{\text{pseudo}},
\end{equation}
\noindent where \(\mathcal{L}_{KL} = \text{KL}\left(W_0 h_M / \tau, q / \tau \right)\) denotes the KL divergence loss, which is used to guide the distillation of task-relevant knowledge from LLMs and \(\mathcal{L}_{\text{pseudo}}\) is the loss associated with the pseudo-labels, given by \(\mathcal{L}_{\text{pseudo}} = -\tilde{\mathbf{y}} \log(\text{MLP}(\hat{h}_\text{LLM}))\). This stage of training enables the model to learn task-relevant semantic information under the guidance of LLM task-relevant knowledge, preparing the model for adapting to downstream tasks. Therefore, during the downstream task fine-tuning stage, Our model is capable of rapidly generalizing to downstream tasks with the guidance of limited labeled data. The loss is computed as:
\begin{equation}
\mathcal{L}_{\text{pre}} = \mathcal{L}_{KL} + \mathcal{L}_{\text{true}},
\end{equation}
where \(\mathcal{L}_{\text{true}}\) depends on the type of task:
\begin{equation}
\mathcal{L}_{\text{true}} =
\begin{cases} 
- y \log(\hat{y}), & \text{for classification}, \\
(y - \hat{y})^2, & \text{for regression}.
\end{cases}
\end{equation}
\noindent This final step refines the model's performance on the specific task using the limited number of labeled examples available.

\section{Experiment} 
In this section, we first introduce the setup for few-shot tabular data prediction and provide the implementation details. We then evaluate the proposed model against with baseline, including traditional machine learning methods, few-shot tabular algorithms, and LLM-based frameworks. Additionally, we conduct extensive ablation studies to test the effectiveness of the proposed modules.






\subsection{Experimental Settings} 
\begin{table}[t]
  \centering
   \resizebox{0.4\textwidth}{!}{
    \begin{tabular}{lccc}
    \toprule
    Data  & \# of samples & \# of cat/num feature  & Task type \\
    \midrule
    Bank  & 45211 & 8/8    & classification \\
    Blood & 748   & 0/4    & classification \\
    Credit-g & 1000  & 12/8     & classification \\
    Diabetes & 768   & 0/8     & classification \\
    Heart & 918   & 4/7    & classification \\
    Myocardial & 1700  & 94/17    & classification \\
    Abalone & 4177  & 1/7     & regression \\
    Boston & 506   & 2/11   & regression \\
    Cholesterol & 303   & 9/4     & regression \\
    \bottomrule
    \end{tabular}%
    }
  \caption{Basic information of each dataset used in
our experiments.}
  \label{tab:datasets}%
\end{table}%

\begin{table*}[t]
  \centering
  \resizebox{1\textwidth}{!}{
    \begin{tabular}{cccccccccccccc}
    \toprule
    data  & shot  & LogReg & XGBoost & RandomForest & SCARF & TabPFN & STUNT & In-context & TABLET & TabLLM & FeatLLM & \ours \\
    \midrule
\multirow{5}[0]{*}{Heart $\uparrow$} & 4     & 70.54\(_{3.83}\) & 50.00\(_{0.00}\) & 70.85\(_{2.02}\) & 59.38\(_{3.42}\) & 67.33\(_{15.29}\) & \textbf{88.27\(_{3.32}\)} & 60.76\(_{4.00}\) & 68.19\(_{11.17}\) & 59.74\(_{4.49}\) & 75.66\(_{4.59}\) & \underline{86.10\(_{5.42}\)} \\
          & 8     & 78.12\(_{10.59}\) & 55.88\(_{3.98}\) & 79.43\(_{4.28}\) & 74.35\(_{6.93}\) & 77.89\(_{2.34}\) & \underline{88.78\(_{2.38}\)} & 65.46\(_{3.77}\) & 69.85\(_{10.82}\) & 70.14\(_{7.91}\) & 79.46\(_{2.16}\) & \textbf{89.59\(_{4.09}\)} \\
          & 16    & 83.02\(_{3.70}\) & 78.62\(_{7.14}\) & 83.45\(_{3.95}\) & 83.66\(_{5.91}\) & 81.45\(_{5.05}\) & \underline{89.13\(_{2.10}\)} & 67.00\(_{7.83}\) & 68.39\(_{11.73}\) & 81.72\(_{3.92}\) & 83.71\(_{1.88}\) & \textbf{92.10\(_{2.79}\)} \\
          & 32    & 84.84\(_{3.53}\) & 87.11\(_{1.22}\) & 88.77\(_{2.36}\) & 88.45\(_{2.43}\) & 88.00\(_{2.34}\) & \underline{89.65\(_{3.04}\)} & 71.94\(_{3.88}\) & 71.90\(_{9.07}\) & 87.43\(_{2.32}\) & 87.19\(_{3.66}\) & \textbf{93.00\(_{2.68}\)} \\
          & 64    & 89.74\(_{3.30}\) & 89.99\(_{3.82}\) & 89.74\(_{2.62}\) & \underline{90.97\(_{1.55}\)} & 90.20\(_{2.57}\) & 89.62\(_{3.16}\) & OOW   & OOW   & 89.78\(_{2.59}\) & 88.08\(_{4.11}\) & \textbf{93.27\(_{2.33}\)} \\\midrule
\multirow{5}[0]{*}{Myocardial $\uparrow$} & 4     & 51.25\(_{3.85}\) & 50.00\(_{0.00}\) & 51.91\(_{4.49}\) & 47.70\(_{4.10}\) & OOW  & 52.77\(_{2.01}\) & OOW   & OOW  & OOW   & \underline{52.87\(_{3.44}\)} & \textbf{55.12\(_{2.86}\)} \\
          & 8     & 55.34\(_{1.11}\) & 55.63\(_{2.92}\) & 52.77\(_{5.83}\) & 49.37\(_{3.41}\) & OOW  & 55.40\(_{4.41}\) & OOW   & OOW   & OOW  & \underline{56.22\(_{1.64}\)} & \textbf{59.85\(_{3.25}\)} \\
          & 16    & 60.00\(_{5.16}\) & 56.55\(_{12.22}\) & 54.16\(_{4.53}\) & 54.31\(_{1.42}\) & OOW   & \underline{61.22\(_{3.45}\)} & OOW   & OOW   & OOW   & 55.32\(_{9.15}\) & \textbf{61.96\(_{4.49}\)} \\
          & 32    & 58.63\(_{0.96}\) & 57.31\(_{5.31}\) & 46.43\(_{5.02}\) & 53.52\(_{0.74}\) & OOW   & \underline{60.76\(_{1.58}\)} & OOW   & OOW   & OOW   & 60.02\(_{4.02}\) & \textbf{62.40\(_{6.40}\)} \\
          & 64    & 57.04\(_{1.94}\) & 56.18\(_{2.85}\) & 55.00\(_{4.73}\) & 54.41\(_{2.00}\) & OOW   & 59.79\(_{0.56}\) & OOW   & OOW   & OOW   & \underline{61.47\(_{3.91}\)} & \textbf{63.84\(_{1.76}\)} \\ \midrule
\multirow{5}[1]{*}{Bank $\uparrow$} & 4     & 63.70\(_{3.87}\) & 50.00\(_{0.00}\) & 60.59\(_{3.90}\) & 58.53\(_{5.49}\) & 63.19\(_{11.60}\) & 56.34\(_{12.82}\) & 61.38\(_{1.30}\) & 58.11\(_{6.29}\) & 62.51\(_{8.95}\) & \underline{70.45\(_{3.69}\)} & \textbf{74.33\(_{2.57}\)} \\
          & 8     & 72.52\(_{3.21}\) & 58.78\(_{10.54}\) & 61.74\(_{9.91}\) & 55.28\(_{11.88}\) & 62.81\(_{7.84}\) & 63.01\(_{8.78}\) & 69.57\(_{13.35}\) & 69.08\(_{6.00}\) & 63.19\(_{5.79}\) & \underline{75.85\(_{6.66}\)} & \textbf{79.91\(_{2.88}\)} \\
          & 16    & 77.51\(_{3.09}\) & 70.34\(_{5.86}\) & 65.67\(_{10.43}\) & 65.81\(_{1.79}\) & 73.79\(_{2.21}\) & 69.85\(_{0.95}\) & 69.76\(_{8.55}\) & 69.40\(_{11.28}\) & 63.73\(_{6.43}\) & \underline{78.41\(_{1.08}\)} & \textbf{82.26\(_{3.56}\)} \\
          & 32    & \underline{79.63\(_{3.57}\)} & 76.25\(_{1.26}\) & 74.29\(_{5.32}\) & 68.45\(_{1.06}\) & 77.71\(_{3.56}\) & 71.64\(_{1.65}\) & 66.93\(_{5.67}\) & 73.61\(_{9.28}\) & 66.51\(_{3.92}\) & 78.37\(_{4.50}\) & \textbf{83.23\(_{9.26}\)} \\
          & 64    & \underline{82.27\(_{1.61}\)} & 81.92\(_{1.00}\) & 79.55\(_{3.19}\) & 68.28\(_{3.97}\) & 82.14\(_{2.28}\) & 72.26\(_{1.62}\) & OOW   & OOW   & 70.83\(_{3.43}\) & 81.18\(_{6.17}\) & \textbf{85.17\(_{7.71}\)} \\ \midrule
\multirow{5}[0]{*}{Boold $\uparrow$} & 4     & 56.79\(_{26.02}\) & 50.00\(_{0.00}\) & 48.50\(_{12.82}\) & 56.22\(_{21.00}\) & 58.72\(_{19.16}\) & 48.57\(_{6.04}\) & 56.30\(_{12.43}\) & 56.45\(_{15.45}\) & 55.87\(_{13.49}\) & \underline{68.34\(_{7.48}\)} & \textbf{70.75\(_{3.64}\)} \\
          & 8     & 68.51\(_{5.16}\) & 59.97\(_{1.36}\) & 63.43\(_{11.03}\) & 65.77\(_{5.00}\) & 66.30\(_{10.01}\) & 60.00\(_{4.84}\) & 58.99\(_{10.12}\) & 56.37\(_{11.56}\) & 66.01\(_{9.25}\) & \textbf{70.37\(_{3.23}\)} & \underline{69.97\(_{3.25}\)} \\
          & 16    & 68.30\(_{6.16}\) & 63.28\(_{7.62}\) & 65.98\(_{6.49}\) & 66.27\(_{5.04}\) & 64.14\(_{6.80}\) & 54.76\(_{4.53}\) & 56.59\(_{5.21}\) & 60.62\(_{4.13}\) & 65.14\(_{7.55}\) & \underline{70.07\(_{5.19}\)} & \textbf{72.46\(_{1.87}\)} \\
          & 32    & 67.39\(_{4.46}\) & 66.41\(_{6.37}\) & 63.46\(_{4.43}\) & 69.71\(_{6.24}\) & 68.65\(_{4.37}\) & 59.87\(_{3.72}\) & 58.69\(_{1.53}\) & 57.94\(_{4.16}\) & 69.95\(_{3.39}\) & \underline{71.13\(_{4.38}\)} & \textbf{74.81\(_{2.67}\)} \\
          & 64    & 71.76\(_{2.56}\) & 69.46\(_{2.96}\) & 68.83\(_{5.61}\) & 72.75\(_{4.36}\) & \underline{73.88\(_{1.97}\)} & 61.75\(_{2.19}\) & 65.79\(_{2.05}\) & 63.47\(_{7.36}\) & 70.88\(_{1.58}\) & 71.04\(_{4.36}\) & \textbf{74.73\(_{2.08}\)} \\ \midrule
\multirow{5}[0]{*}{Diabetes $\uparrow$} & 4     & 57.09\(_{18.84}\) & 50.00\(_{0.00}\) & 52.50\(_{7.77}\) & 62.35\(_{7.48}\) & 56.28\(_{13.01}\) & 64.22\(_{6.78}\) & 71.71\(_{5.31}\) & 63.96\(_{3.32}\) & 70.42\(_{3.69}\) & \textbf{80.28\(_{0.75}\)} & \underline{72.06\(_{5.04}\)} \\
          & 8     & 65.52\(_{13.18}\) & 50.86\(_{22.03}\) & 65.34\(_{8.84}\) & 64.69\(_{13.33}\) & 69.08\(_{9.68}\) & 67.39\(_{12.92}\) & 72.21\(_{2.07}\) & 65.47\(_{3.95}\) & 64.30\(_{5.88}\) & \textbf{79.38\(_{1.66}\)} & \underline{73.70\(_{4.94}\)} \\
          & 16    & 73.44\(_{0.52}\) & 65.69\(_{6.54}\) & 65.69\(_{6.33}\) & 71.86\(_{3.16}\) & 73.69\(_{3.21}\) & 73.79\(_{6.48}\) & 71.64\(_{5.05}\) & 66.71\(_{0.76}\) & 67.34\(_{2.79}\) & \textbf{80.15\(_{1.35}\)} & \underline{76.78\(_{4.69}\)} \\
          & 32    & 73.95\(_{3.32}\) & 72.97\(_{3.77}\) & 71.27\(_{8.04}\) & 72.91\(_{3.09}\) & 75.22\(_{3.21}\) & 76.70\(_{4.55}\) & 73.32\(_{1.59}\) & 66.97\(_{1.75}\) & 69.74\(_{4.41}\) & \textbf{80.06\(_{1.18}\)} & \underline{77.01\(_{2.44}\)} \\
          & 64    & 74.52\(_{1.59}\) & 72.56\(_{3.17}\) & 76.92\(_{2.39}\) & 74.44\(_{4.13}\) & 77.82\(_{3.49}\) & \underline{78.64\(_{3.32}\)} & 70.22\(_{4.09}\) & 69.27\(_{6.15}\) & 71.56\(_{4.55}\) & \textbf{80.91\(_{1.62}\)} & 78.32\(_{2.49}\) \\ \midrule
          \multirow{5}[0]{*}{Credit-g $\uparrow$} & 4     & 52.68\(_{4.46}\) & 50.00\(_{0.00}\) & \underline{57.00\(_{10.75}\)} & 48.92\(_{4.60}\) & 54.00\(_{7.34}\) & 48.80\(_{6.76}\) & 52.99\(_{4.08}\) & 54.33\(_{6.54}\) & 51.90\(_{9.40}\) & 55.94\(_{1.10}\) & \textbf{58.21\(_{0.93}\)} \\
          & 8     & 55.52\(_{8.88}\) & 52.22\(_{4.90}\) & \underline{59.84\(_{7.33}\)} & 55.26\(_{3.92}\) & 52.58\(_{11.27}\) & 54.50\(_{8.25}\) & 52.43\(_{4.36}\) & 52.90\(_{5.79}\) & 56.42\(_{12.89}\) & 57.42\(_{3.10}\) & \textbf{62.90\(_{4.52}\)} \\
          & 16    & 58.26\(_{5.17}\) & 56.23\(_{4.37}\) & 58.42\(_{8.36}\) & 59.22\(_{11.38}\) & 58.91\(_{8.04}\) & 57.63\(_{7.58}\) & 55.29\(_{4.80}\) & 51.65\(_{4.02}\) & \underline{60.38\(_{14.03}\)} & 56.60\(_{2.22}\) & \textbf{67.11\(_{1.22}\)} \\
          & 32    & 67.85\(_{5.78}\) & 65.33\(_{6.28}\) & 57.48\(_{5.73}\) & \underline{72.60\(_{7.18}\)} & 66.27\(_{5.06}\) & 63.24\(_{5.47}\) & OOW   & OOW   & 68.64\(_{3.86}\) & 61.79\(_{10.25}\) & \textbf{73.13\(_{1.49}\)} \\
          & 64    & 72.77\(_{9.29}\) & 70.79\(_{2.34}\) & 68.53\(_{4.99}\) & \underline{73.12\(_{6.23}\)} & 68.95\(_{6.14}\) & 70.97\(_{4.95}\) & OOW   & OOW   & 70.80\(_{4.09}\) & 66.43\(_{2.90}\) & \textbf{75.65\(_{1.05}\)} \\ \midrule
    \end{tabular}%
    }
   \caption{Evaluation results, including the AUC scores across six classification datasets. The best performances are highlighted in bold, and second-best are underlined. For consistency and clarity, all subsequent tables in this paper adhere to the same format: metric values are averaged over three random seeds, with standard deviation provided after it. Out of In-context Window (OOW) indicates that the input length exceeds the maximum context window limit of the model and cannot be processed.}
  \label{tab:classification}%
\end{table*}

\noindent \textbf{Dataset.} As shown in Table \ref{tab:datasets}, we evaluate the proposed method using nine real-world datasets, including six classification tasks and three regression tasks. Each dataset provides metadata such as a clear description for each attribute and task. Refer to Appendix A for details of the dataset 

\noindent \textbf{Baselines.} We compare the proposed method against ten baseline models. The first is conventional tabular learning approaches: (1) Logistic Regression (LogReg)~\cite{lavalley2008logistic}, (2) XGBoost~\cite{chen2016xgboost}, and (3) Random Forest~\cite{breiman2001random}. The next two baselines leverage unlabeled datasets: (4) SCARF~\cite{bahri2021scarf} and (5) STUNT~\cite{nam2023stunt}, though it is worth noting that in real-world scenarios while obtaining unlabeled data is typically straightforward, the process of labeling it can be expensive and time-consuming. Another recent baseline, (6) TabPFN~\cite{hollmann2022tabpfn}, generates synthetic datasets with diverse distributions to pretrain the model. The final group of baselines utilizes LLMs: (7) In-context learning~\cite{wei2022emergent}, which embeds few-shot training samples directly into the input prompt without fine-tuning; (8) TABLET~\cite{slack2023tablet}, which enriches the prompt with additional hints through an external classifier using rule sets and prototypes to improve inference quality; (9) TabLLM~\cite{hegselmann2023tabllm}, which applies parameter-efficient tuning techniques such as IA3 to train the LLMs with few-shot samples; and (10) FeatLLM~\cite{han2024large}, which treats the LLMs as a feature engineer, automating feature extraction and using the engineered features for few-shot tabular classification.

\noindent \textbf{Implementation details.} We obtain task-relevant knowledge from the LLaMA2 series models~\cite{touvron2023llama}. The hidden layer dimensions of the Feed Forward Network (FFN) and other components are set to 256 and 128, respectively. For the Semantic-aware Tabular Encoder, we use a 2-layer, 8-head transformer, while the knowledge adapter consists of a 2-layer, 2-head transformer. Dropout is set to 0.1.  During the meta-learning phase, The Adam optimizer is used for training, with the learning rate set to 1e-4. The activation vector in the LLMs is obtained from the 30 layers.  For the fine-tuning phase, the learning rate is reduced to 1e-5 due to the limited number of labeled samples. For evaluation, we use the Area Under the Curve (AUC) metric for classification tasks and Mean Squared Error (MSE) for regression tasks.
\subsection{Results and Analysis}

\textbf{\ours~offers an effective and comprehensive solution for few-shot tabular prediction.}
For \textbf{Effectiveness:} As shown in Table~\ref{tab:classification}, \ours~outperforms or comparable of the current state-of-the-art baseline across all datasets and shot setting. For example, in classification tasks, our method exceeds the performance of the state-of-the-art method FeatLLM in all shot settings, achieving an average performance improvement of 4.22\%. This improvement is primarily due to \ours’s ability to leverage both unlabeled data and task-related knowledge from LLM, which is something that existing methods cannot achieve.

For \textbf{Comprehensiveness:} \ours~ directly applicable to regression tasks without the need for modifications. Detailed experimental results can be found in Appendix B. In contrast, existing few-shot learning methods are often designed specifically for classification tasks and cannot be directly applied to regression. While large language models (LLMs) can handle regression tasks, they tend to produce inaccurate predictions, often due to spurious patterns or ``hallucination.''
\subsection{Ablation Study}
\begin{table}[t]
\renewcommand{\arraystretch}{1.3}
  \centering
\resizebox{0.48\textwidth}{!}{
    \begin{tabular}{ccccccccc}
    \toprule
    SaTE & LLM   & Meta & Size & 4     & 8     & 16    & 32    & 64 \\
    \midrule
          \ding{51}     &&       &  & 85.16\(_{4.44}\) & 84.86\(_{6.73}\) & 88.59\(_{8.54}\) & 90.03\(_{3.60}\) & 91.07\(_{2.75}\) \\
          \ding{51}     && \ding{51}     &       & 81.27\(_{10.13}\) & 86.26\(_{7.29}\) & 90.43\(_{5.07}\) & 90.52\(_{2.59}\) & 91.99\(_{2.46}\) \\
    \ding{51}     & \ding{51}     &       & 13B      & 84.64\(_{4.16}\) & 84.71\(_{7.93}\) & 89.63\(_{6.21}\) & 90.63\(_{4.90}\) & 91.52\(_{4.53}\) \\
    & \ding{51}     & \ding{51}     &       13B & 76.57\(_{6.56}\) & 83.13\(_{1.79}\) & 86.52\(_{2.34}\) & 83.09\(_{5.95}\) & 89.68\(_{5.81}\) \\
    \ding{51}     & \ding{51}     & \ding{51}     & 7B    & 85.48\(_{5.85}\) & 87.39\(_{6.01}\) & 92.08\(_{3.24}\) & 92.93\(_{2.71}\) & 93.13\(_{2.60}\) \\
     \midrule
     \ding{51}     & \ding{51}     & \ding{51}     &       13B& \textbf{86.10\(_{5.42}\)} & \textbf{89.59\(_{4.09}\)} & \textbf{92.10\(_{2.79}\)} & \textbf{93.00\(_{2.68}\)} & \textbf{93.27\(_{2.33}\)} \\
   
    \bottomrule

    \end{tabular}%
    }
  \caption{Ablation experiments conducted on Heart dataset}
  \label{tab:ablation}%
  \vspace{-17pt}
\end{table}%
 \textbf{Each module in \ours~plays a crucial role.} We conduct ablation studies to assess the contribution of key components to the model's performance. Specifically, we evaluate four factors: (1) \textbf{SaTE} explores the decision of whether to use a semantic-aware table encoder. If a semantic-aware approach is not employed, a non-semantic-aware encoder from STUNT~\cite{nam2023stunt} is used as an alternative, (2) \textbf{LLM} indicates whether task-relevant knowledge is extracted from the LLMs, (3) \textbf{Meta} refers to the process of applying meta-learning techniques to unlabeled data, and (4) \textbf{Size} represents the effect of varying LLM scales on performance.

Table \ref{tab:ablation} presents the impact of various ablation experiments on the AUC across the Heart datasets. In all cases, modifying any of the ablated components leads to a decline in performance. Specifically, we replaced \textbf{SaTE} with the encoder from STUNT~\cite{nam2023stunt} and the result  reveal a maximum performance drop, particularly with few-shot labeled samples. For instance, with just 4 labeled samples, performance decreased by nearly 10\%. This highlights the critical role of capturing the unique properties of tabular data and the importance of incorporating semantic information from column names, especially when working with limited labeled data. Besides, without extracting task-relevant knowledge from the LLMs using metadata, or meta-learning with unlabeled data, also decrease the model's performance. This highlights the importance of learning task-related semantic knowledge in the meta-learning process. Interestingly, increasing the size of the selected LLMs does not result in significant performance gains, possibly due to the 7B model has already contained all the necessary task-related semantic knowledge. 

\subsection{Impact of Labeled Sample Size and LLM Activation Layers on \ours}
\label{sec:RQ3}
\begin{figure*}[t]
\centering
\includegraphics[width=1\textwidth]{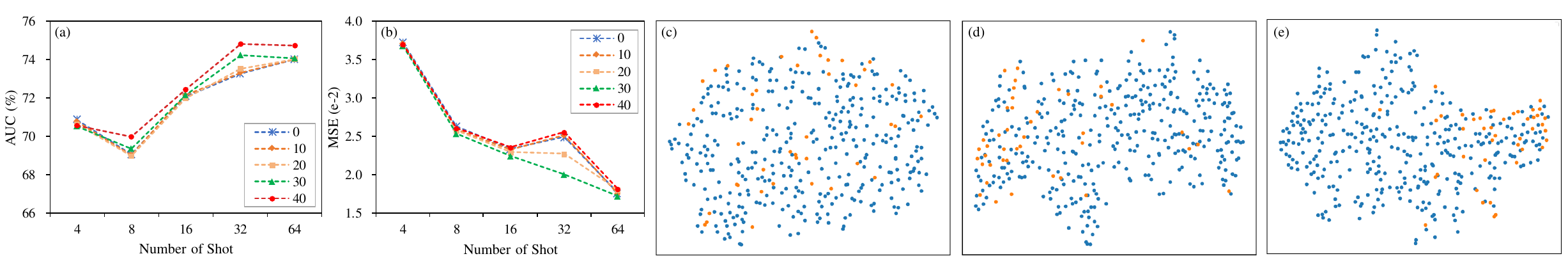}
\caption{\label{figure:p_l}\ours's parameter experiments and representation visualizations. (a-b) Impact of the number of labeled samples and LLM activation layers on \ours's performance in classification and regression tasks. (c-e) Visualization of \ours's learned representations: (c) original representations, and \ours~representations trained on (d) 4 and (e) 64 labeled samples.}
\vspace{-12pt}
\end{figure*}


\textbf{The deeper of LLM, the more sufficient knowledge it contains.}  Fig.~\ref{figure:p_l} (a-b) illustrates the effect of activation in different layers \( L \) of the LLM and the number of labeled samples on model performance.  Fig.~\ref{figure:p_l} (a) presents AUC values for classification tasks on the Blood dataset, while  Fig.~\ref{figure:p_l} (b) shows MSE values for regression tasks on the Abalone dataset. As the number of labeled samples increases, model performance improves across both tasks. In classification tasks, performance improves as deeper layers of the LLM are activated. This suggests that higher-order knowledge is accumulated in the later layers. In contrast, for regression tasks, optimal performance is achieved at around 30 layers. In summary, adding more layers may not improve the model's performance and could instead introduce noise, leading to a decline in the quality of learned representations.
\subsection{What Latent Knowledge Does \ours~Transfer From the LLM}

\begin{figure}[t]
\centering
\includegraphics[width=0.5\textwidth]{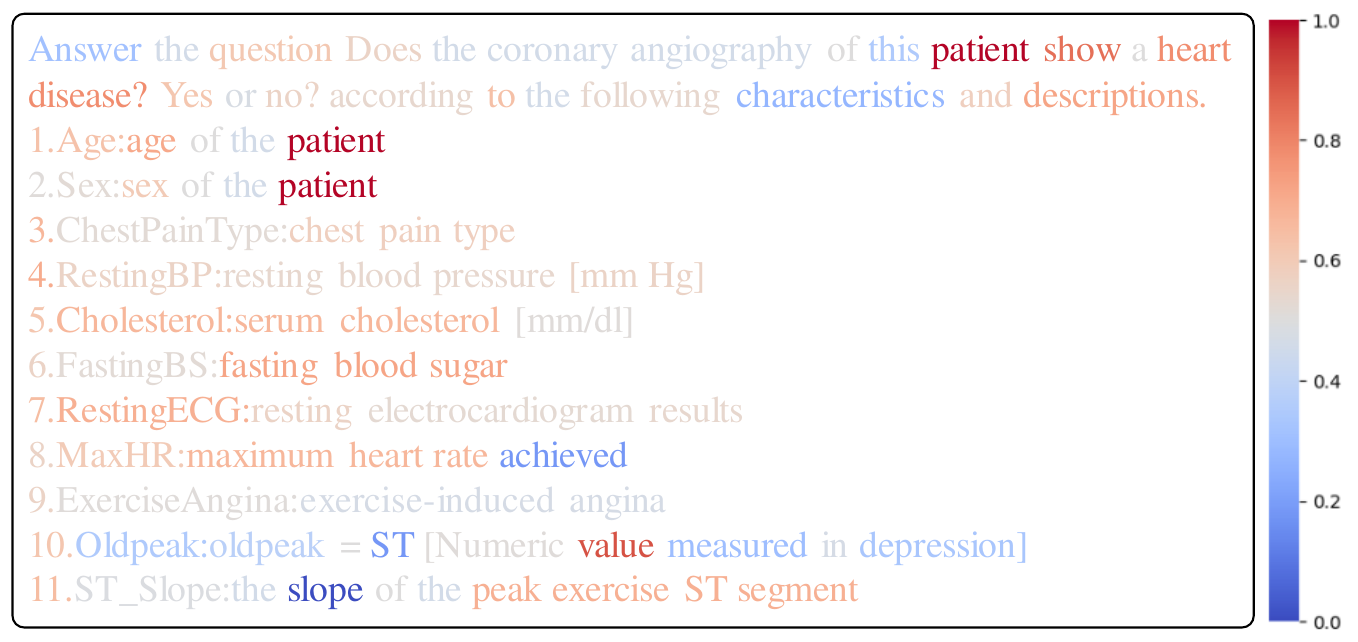}
\caption{\label{figure:case_study}The heatmap visualizes the task-related semantic information learned by the model during the meta-learning stage. Higher values represent a greater semantic similarity between the model's learned representations and the semantic knowledge associated with the LLM task.}
\vspace{-10pt}
\end{figure}
\textbf{\ours~can retrieve task-related semantic information from metadata.} To examine the task-related semantic information embedded in the representations learned by \myVar \ during the unlabeled data meta-learning stage, we measured the semantic similarity between the learned representations and the LLM activation at the word level using metadata. This was visualized through a heatmap, as shown in  Fig. \ref{figure:case_study}. Higher values, indicated by redder colors, reflect greater semantic similarity. As an example of a downstream task, this involves predicting whether a patient has heart disease. The results of the heatmap demonstrate that our model, after meta-learning on unlabeled data, captures task-relevant semantic information. For example, the representations learned by \ours~exhibit high semantic similarity with ``patient'' and ``heart disease,'' indicating its effective learning of task-relevant semantic knowledge. Furthermore, the model identifies and emphasizes key features such as ``age,'' ``cholesterol,'' ``fast blood sugar,'' ``maximum heart rate,'' and ``peak exercise ST segment,'' which are crucial for determining heart disease. 
\subsection{Comparison of LLM Invocation Costs}
\begin{table}[t]
  \centering
  \resizebox{0.45\textwidth}{!}{
    \begin{tabular}{cccccc}
    \toprule
    \multirow{2}[4]{*}{Stage} & \multicolumn{5}{c}{Methods} \\
\cmidrule{2-6}          & In-context & TABLET & TabLLM & FeatLLM & \ours \\
    \midrule
    Preprocessing & 0     & 0     & 0     & 0     & 1 \\
    Training & 0     & 1     & 1     & 30    & 0 \\
    Inference & 9043  & 9043  & 9043  & 0     & 0 \\
    \midrule
    Totle & 9043  & 9044  & 9044  & 30    & 1 \\
    \bottomrule
    \end{tabular}%
    }
      \caption{Compare our method with the LLM baseline in terms of the number of LLM calls made during the preprocessing, training, and inference stages.}
  \label{tab:LLM_call}%
\vspace{-17pt}
\end{table}%
\textbf{\ours~has an efficiency advantage in calling LLM.} Table \ref{tab:LLM_call} compares the number of LLM calls required by \ours and other LLM-based baselines across the data preprocessing, training, and inference stages. Specifically, TABLET, In-context and TabLLM require calling the LLM once for each test sample during inference, resulting in significant overhead. While FeatLLM limits LLM calls to the training phase to assist in learning, the number of calls still scales with the size of the training dataset. Our method calls the LLM only once during preprocessing to gather task-relevant knowledge, which is then stored locally. As a result, our approach significantly reduces the overall number of LLM calls throughout the entire process.

\subsection{Visualization Experiments}

\textbf{\ours~can generate high-quality representations with limited labeled data.} The visualization of representations on the Bank dataset is shown in  Fig. \ref{figure:p_l} (c-e), where the original representations refer to those generated without training. \ours~learns effective representations in few-shot settings. With just four labeled samples, it produces a separable representation, and this separability improves as the number of labeled samples increases to 64. This demonstrates the model's strong ability to learn useful features with minimal supervision.
\vspace{-12pt}
\section{Conclusion}
This paper introduces a novel framework, \ours, aimed at addressing the challenges of few-shot tabular learning by extracting latent-level knowledge from large language models (LLMs) during the training process. \ours~positions the LLM as a ``teacher'' that provides task-relevant guidance to the downstream tabular model and enable a knowledge-guided training process that mitigates overfitting, a common issue in few-shot settings, and promotes convergence toward a generalized model. Additionally, \ours~incorporates an unsupervised pre-training stage that leverages unlabeled data for robust parameter initialization, further enhancing the framework's performance. Extensive experiments demonstrate the effectiveness of \ours~on a variety of few-shot tabular learning tasks. The results show that \ours~consistently outperforms existing methods across multiple datasets and prediction tasks. Ablation studies further validate the importance of latent-level knowledge over text-level knowledge and highlight the contribution of each component to the overall success of the framework.
\section*{Acknowledgments} This work was supported by a grant from the National Natural Science Foundation of China under grants (No.62372211,  62272191), and the Science and Technology Development Program of Jilin Province (No.20250102216JC), and the International Science and Technology Cooperation Program of Jilin Province (No.20230402076GH, No. 20240402067GH), and Reform Commission Foundation of Jilin Province (No.2024C003).
\appendix

\bibliographystyle{named}
\bibliography{ijcai25}

\end{document}


\maketitle
\appendix

\section{Dataset Details}

We evaluate the proposed method using nine real-world datasets, including six classification tasks and three regression tasks. The classification datasets are as follows:
\begin{itemize}
    \item \textit{Bank}~\cite{moro2014data}, which predicts whether a customer will subscribe to a term deposit;

    \item \textit{Blood}~\cite{yeh2009knowledge}, which predicts whether donors will return for subsequent donations;

    \item \textit{Credit-g}~\cite{kadra2021well}, which predicts whether an individual poses a good or bad credit risk;

    \item \textit{Diabetes}~\footnote{\url{kaggle.com/datasets/uciml/pima-indians-diabetes-database}}, which predicts the presence of diabetes in an individual;

    \item \textit{Heart}~\footnote{\url{kaggle.com/datasets/fedesoriano/heart-failure-prediction}}, which predicts the occurrence of coronary artery disease;

    \item \textit{Myocardial}~\cite{myocardial_infarction_complications_579}, which predicts whether an individual suffers from chronic heart failure.

\end{itemize}

     Three regression datasets in OpenML~\cite{vanschoren2014openml} include: 
     \begin{itemize}
         \item \textit{Abalone}, which predicts the age of abalone;
         
         \item \textit{Aoston}, which predicts the housing prices in Boston;

         \item \textit{Cholesterol}, which predicts the value of serum cholesterol in mg/dl.
         
     \end{itemize}
     
\section{Regression experiment results}

\begin{table}[ht]
  \centering
  \resizebox{0.5\textwidth}{!}{
    \begin{tabular}{ccccccccc}
    \toprule
    data  & shot  & LogReg & XGBoost & RandomForest & SCARF  & In-context & \ours \\
    \midrule
 
\multirow{5}[0]{*}{Abalone \downarrow}  & 4     & 5.69e-2\(_{3.51\text{e-}2}\) & \underline{3.41e-2\(_{1.17\text{e-}2}\)} & 3.98e-2\(_{8.06\text{e-}3}\) & 4.28e-2\(_{1.70\text{e-}2}\)  & 4.10e-2\(_{2.11\text{e-}2}\)  & \textbf{3.40e-2\(_{2.52\text{e-}2}\)} \\
          & 8     & 3.12e-1\(_{1.40\text{e-}1}\) & 5.63e-2\(_{5.56\text{e-}2}\) & \underline{4.20e-2\(_{3.87\text{e-}3}\)} & 3.46e-1\(_{1.60\text{e-}1}\)  & 7.63e-2\(_{5.90\text{e-}2}\)  & \textbf{3.04e-2\(_{1.41\text{e-}2}\)} \\
          & 16    & 5.29e-2\(_{3.54\text{e-}2}\) & \underline{2.75e-2\(_{6.65\text{e-}3}\)} & 4.08e-2\(_{8.11\text{e-}3}\) & 6.23e-2\(_{4.65\text{e-}2}\)  & 7.54e-2\(_{3.70\text{e-}2}\)  & \textbf{2.60e-2\(_{5.00\text{e-}3}\)} \\
          & 32    & 2.61e-2\(_{2.37\text{e-}2}\) & 2.37e-2\(_{4.57\text{e-}3}\) & 2.31e-2\(_{4.78\text{e-}3}\) & \textbf{1.70e-2\(_{5.18\text{e-}3}\)}  & 9.15e-2\(_{1.09\text{e-}2}\)   & \underline{2.17e-2\(_{2.94\text{e-}3}\)} \\
          & 64    & 2.60e-2\(_{2.34\text{e-}2}\) & 2.16e-2\(_{3.10\text{e-}3}\) & 1.86e-2\(_{8.48\text{e-}3}\) & \underline{1.69e-2\(_{5.08\text{e-}3}\)} & 9.49e-2\(_{8.71\text{e-}3}\)   & \textbf{1.65e-2\(_{4.91\text{e-}3}\)} \\\midrule

\multirow{5}[0]{*}{Boston \downarrow} & 4     & 3.00e-1\(_{2.88\text{e-}1}\) & 7.08e-2\(_{3.63\text{e-}2}\) & \underline{5.31e-2\(_{7.76\text{e-}3}\)} & 1.09e-1\(_{1.21\text{e-}1}\)  & 1.05e-1\(_{7.76\text{e-}3}\)  & \textbf{3.68e-2\(_{2.72\text{e-}3}\)} \\
          & 8     & 4.16e-2\(_{2.13\text{e-}2}\) & 5.56e-2\(_{7.66\text{e-}3}\) & \underline{3.81e-2\(_{6.59\text{e-}3}\)} & 4.26e-2\(_{9.83\text{e-}3}\)  & 4.84e-1\(_{6.59\text{e-}3}\)  & \textbf{2.54e-2\(_{8.39\text{e-}3}\)} \\
          & 16    & 1.06e-1\(_{6.25\text{e-}2}\) & 4.16e-2\(_{1.70\text{e-}2}\) & \underline{2.82e-2\(_{1.66\text{e-}2}\)} & 1.06e-1\(_{8.10\text{e-}2}\) & 2.67e-1\(_{1.66\text{e-}2}\)  & \textbf{2.23e-2\(_{1.04\text{e-}2}\)} \\
          & 32    & 3.22e-2\(_{9.69\text{e-}3}\)& 4.23e-2\(_{1.11\text{e-}2}\) & \underline{2.98e-2\(_{1.62\text{e-}2}\)} & 3.19e-2\(_{9.75\text{e-}3}\) & 3.94e-1\(_{1.62\text{e-}2}\)   & \textbf{2.00e-2\(_{6.66\text{e-}3}\)} \\
          & 64    & 2.18e-2\(_{3.52\text{e-}3}\) & 2.23e-2\(_{3.23\text{e-}2}\) & \underline{1.86e-2\(_{2.66\text{e-}3}\)} & 2.17e-2\(_{3.13\text{e-}3}\)  & 2.49e-1\(_{2.66\text{e-}3}\)    & \textbf{1.80e-2\(_{6.38\text{e-}3}\)} \\\midrule

\multirow{5}[0]{*}{Cholesterol \downarrow} & 4     & 1.38e-1\(_{9.71\text{e-}2}\) & 7.79e-2\(_{3.87\text{e-}2}\) & \underline{6.06e-2\(_{3.48\text{e-}2}\)} & 8.62e-2\(_{4.12\text{e-}2}\)  & 1.25e-1\(_{3.48\text{e-}2}\) & \textbf{5.44e-2\(_{2.71\text{e-}2}\)} \\
          & 8     & 2.64e-1\(_{2.32\text{e-}1}\) & 7.04e-2\(_{2.56\text{e-}2}\) & \textbf{4.02e-2\(_{1.38\text{e-}2}\)} & 2.91e-1\(_{3.44\text{e-}1}\)  & 8.74e-2\(_{1.39\text{e-}2}\) & \underline{4.41e-2\(_{8.10\text{e-}3}\)} \\
          & 16    & 11.42\(_{14.04}\) & 4.03e-2\(_{9.95\text{e-}3}\) & \underline{2.42e-2\(_{2.76\text{e-}3}\)} & 11.31\(_{17.77}\)  & 1.22e-1\(_{2.76\text{e-}3}\)  & \textbf{2.26e-2\(_{1.17\text{e-}2}\)} \\
          & 32    & 1.84e-1\(_{1.17\text{e-}1}\) & 3.62e-2\(_{1.62\text{e-}2}\) & \underline{2.31e-2\(_{4.78\text{e-}3}\)} & 1.84e-1\(_{1.57\text{e-}1}\)  & 2.42e-1\(_{4.78\text{e-}3}\)   & \textbf{2.11e-2\(_{5.30\text{e-}3}\)} \\
          & 64    & 2.36e-2\(_{1.79\text{e-}2}\) & 3.48e-2\(_{3.48\text{e-}3}\) & \underline{1.81e-2\(_{5.09\text{e-}3}\)} & 2.37e-2\(_{2.65\text{e-}3}\) & 2.27e-1\(_{5.09\text{e-}3}\)    & \textbf{1.45e-2\(_{3.58\text{e-}3}\)} \\\midrule

    \end{tabular}%
    }
   \caption{Evaluation results, including the MSE scores across three regression datasets. The best performances are highlighted in bold, and second-best are underlined. "-" indicates that this method is specifically designed for classification tasks and cannot handle regression tasks.}
  \label{tab:regression}%
\end{table}

In this section, we compare the performance of \ours~with several baseline methods on three regression datasets. The baseline methods—TabPFN, STUNT, TABLET, TabLLM, and FeatLLM—were originally designed for classification tasks and are not directly applicable to regression tasks. \ours~consistently outperforms these baselines in regression settings. This suggests that the representations learned by \ours, which leverage unlabeled data and LLM-derived task-specific semantic knowledge, are beneficial for both classification and regression tasks. These results highlight the versatility and generality of \ours, demonstrating its capability to effectively handle a wide range of predictive tasks. At the same time, we have discovered some interesting phenomena: although directly prompt large language models (LLMs) can perform regression tasks, they tend to produce inaccurate numerical predictions, often influenced by spurious patterns or ”Hallucination.” Our experiments reveal a notable distinction in LLM performance between classification and regression tasks. Specifically, as the number of samples increases, the regression performance tends to worsen, rather than improve. This decline is likely due to the continuous nature of the label space in regression tasks, which makes it challenging for LLMs to establish accurate mappings between samples and their corresponding labels. Increasing the number of examples not only does not promote the establishment of mapping relationships but also introduces more noise, which further exacerbates the model’s hallucination.
\bibliographystyle{named}
\bibliography{ijcai25}